\pdfoutput=1
\documentclass[11pt]{article}

\usepackage[final]{acl}

\usepackage{times}
\usepackage{latexsym}
\usepackage{annotates}

\usepackage[T1]{fontenc}
\usepackage[utf8]{inputenc}

\usepackage{microtype}

\usepackage{inconsolata}

\usepackage{graphicx}

\usepackage{booktabs}
\usepackage{amsmath}
\usepackage{amsfonts}
\usepackage{adjustbox}
\usepackage{enumitem}
\usepackage{multirow}

\title{Interpretable Text Embeddings and Text Similarity Explanation: A Survey}

\author{Juri Opitz$^{1\ast}$\; Lucas Moeller$^{2\ast}$\; Andrianos Michail$^{1}$\; Sebastian Padó$^{2}$\; Simon Clematide$^{1}$ \medskip\\
  $^1$University of Zurich, Switzerland \\\medskip 
  $^2$University of Stuttgart, Germany \\
  $^1$\texttt{\{jurialexander.opitz,andrianos.michail,simon.clematide\}@uzh.ch} \\  $^2$\texttt{\{lucas.moeller,sebastian.pado\}@ims.uni-stuttgart.de}\\$^\ast$\begin{small}Equal contribution\end{small}}

\begin{document}
\maketitle
\begin{abstract}
Text embeddings are a fundamental component in many NLP tasks, including classification, regression, clustering, and semantic search. 
However, despite their ubiquitous application, challenges persist in interpreting embeddings and explaining similarities between them. 
In this work, we provide a structured overview of methods specializing in inherently interpretable text embeddings and text similarity explanation, an underexplored research area. 
We characterize the main ideas, approaches, and trade-offs. 
We compare means of evaluation, discuss overarching lessons learned and finally identify opportunities and open challenges for future research.
\end{abstract}

\section{Introduction}

Text embedding models \citep{reimers-gurevych-2019-sentence, gao-etal-2021-simcse} are ubiquitous in research and industry, as they promise to map the meaning or content of sentences and documents to useful numerical vector representations (``embeddings''), among which an arithmetic distance (or similarity) can be calculated. Applications range from semantic search and retrieval \citep{ye2016word, GUO2020102067_neural_ir, muennighoff2022sgpt, hambarde2023information, ALATRASH2024100193} to text classification \citep{schopf2022evaluating}, topic modeling \citep{grootendorst2022bertopic}, NLG evaluation \citep{celikyilmaz2020evaluation, nlg_survey, larionov-etal-2023-effeval, chollampatt-etal-2025-cross}, graph reasoning \citep{plenz-etal-2023-similarity}, and retrieval-augmented generation \citep[RAG,][]{NEURIPS2020_6b493230_rag, gao2023retrieval}. Advances in base models \citep{jina, wang-etal-2024-improving-text}, context size \citep{li2023towards}, instruction tuning \citep{allen}, and scalable infrastructure \citep{wang2022text} continuously enhance their capabilities. %\seb{mention LLM2Vec here?} DONE
Most recently, a trend has been to build embedding models from large pre-trained decoders by removing their causal attention masking and continuing to train them contrastively in a Siamese setup using annotated, mined or LLM augmented pairs of similar texts \citep{muennighoff2022sgpt,llm2vec}.
The approach is also widely adopted by the industry \citep{nvembed, gecko, gemini-embed}.
While this shows that knowledge obtained upon generative pretraining can be effectively translated to representation tasks, evidently, a critical component remains contrastive training.
Thus, the learning of informative text representations appears to be closely linked to text similarity.

Yet, with the advancement of text embedding models, a pressing challenge persists: the \textit{interpretability of embeddings} and the \textit{explainability of similarity} derived from them. For instance, when a document is returned in response to a query, we would like to articulate why this document was selected as the most \textit{similar}, or why another was omitted. 
We find interpretability research with a focus on text representation and similarity to be under-represented in the literature.
One reason for this may lie in the pairwise nature of the encountered inputs, which introduces additional complexity:
Similarity depends on interactions between two inputs rather than on features of a single input---a change in one input influences the effect of the second on the prediction \cite{tversky, lin}---and
explanations must account for these interactions.

Importantly, such questions are not just theoretical. 
 In light of laws like the EU AI Act  \citep[``right to explanation'';][]{eu-act}, the demand for transparency is expected to intensify. Thus, there is a strong and timely need for research, overview, and clarity in \textit{all} fields of AI. 
 In this work, we focus on interpretability and explainability in the context of similarity and embedding models.  
 We intend this survey to serve as a resource for researchers interested in these challenges and to lower the entry barrier into this area positioned at the intersection of several research domains, including interpretability, representation learning, and NLP.

\section{Setting the Stage} \label{sec:stage}

We investigate interpretability and explainability in the context of neural text embedding models, and focus on three closely related aspects: (i) the interpretability of the models themselves, (ii) the properties of the text embeddings these models generate, and (iii) the similarity scores derived from comparing such embeddings.

\begin{figure}[t]
    \centering
    \includegraphics[width=\linewidth]{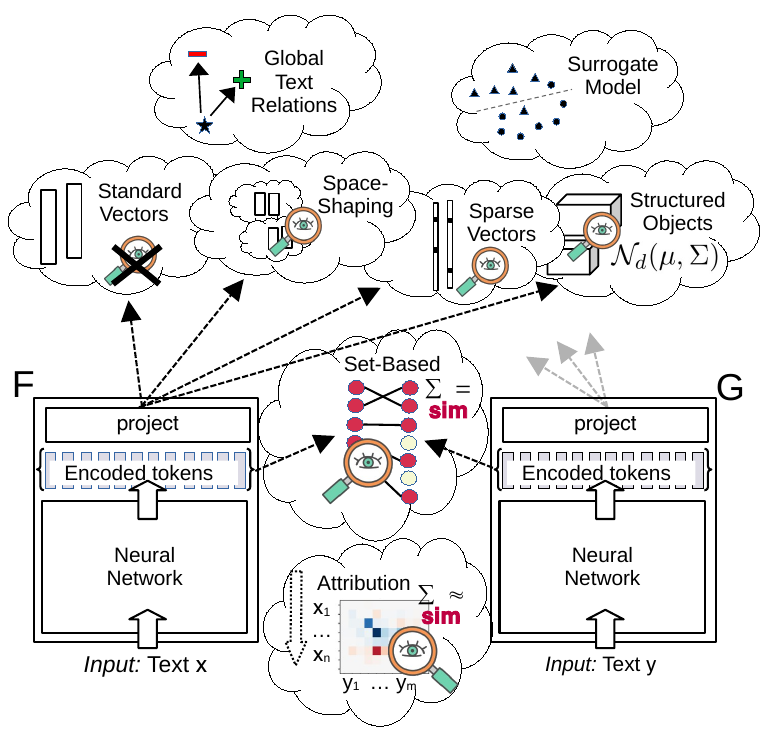}
    \caption{A schema of a standard text encoder architecture with the different interpretable embeddings and explainability approaches, each corresponding to subsections in the text.}
    \label{fig:overview}
\end{figure}

\paragraph{Formal framework (Figure \ref{fig:overview}).} 
Assume two text encoders $F$ and $G$. Their backbone typically consists of a multi-layered neural network.  
In most cases, $F=G$, meaning the networks share weights, a setup known as a \textit{Siamese} network \citep{koch2015siamese}; unless stated otherwise, we refer only to $F$. After the Neural Network has processed and refined the representation of an input text $x$, a last neural layer typically provides a final stage of refined representations, often on the token level basis (encoded tokens). Additionally, in most cases, there is a special last (optional) layer $L+1$ (\textit{project} in Figure \ref{fig:overview}) that projects to a $d$-dimensional space where text representation is independent of text length (standard vectors, Figure~\ref{fig:overview}). This layer often employs averaging or max-pooling  across individual token embedding dimensions, or reproduces the embedding from a specialized token. 

Finally, the similarity of two texts $x, y$ can be efficiently calculated through their embeddings (vectors) $e_x = F(x), e_y = F(y)$ by calculating a function $sim(e_x, e_y) \in \mathbb{R}$. In the simplest case, this can be the dot product $sim(e_x, e_y) = e_x^Te_y$, possibly normalized by length $l_{x,y}=|e_x|_2\cdot|e_y|_2$ to obtain cosine similarity.

So far, the mechanism of text encoding and similarity computation is a standard and ubiquitous procedure. Importantly, this procedure leads to non-interpretable vectors, and consequently yields similarities that escape interpretation or explanation. Next, we elaborate on the parts that allow us to resolve or at least mitigate this issue. Simultaneously, we use these parts in Figure \ref{fig:overview} to establish a scaffold for the structure of this survey.

\paragraph{Paper Structure.} Our formal framework in Figure \ref{fig:overview}  permits us to distinguish between different classes of explainability approaches.  

On the top level, we distinguish between \textbf{interpretable embeddings} (\S \ref{sec:interpret_emb}) and \textbf{post-hoc explanations} (\S \ref{sec:post-hoc}).
We speak of \textit{interpretable} models if the structure of their architecture or embeddings inherently enables insights into their predictions to humans without a need for additional methods or further processing.
\textit{Post-hoc explanations}, on the other hand, generate insights into uninterpretable black-box models by applying an additional method that is not part of the original model's computation.
We divide the former further into \textbf{space-shaping approaches} (\S \ref{sec:space-shaping}) structuring the learned embedding space in interpretable ways, \textbf{sparse representations} (\S \ref{sec:sparse}) yielding human-understandable sparse features, \textbf{structured objects} (\S \ref{sec:geometric}) representing texts as geometric objects instead of simple vectors and \textbf{set-based embeddings} (\S \ref{sec:setbased}) using not a single but multiple vectors to represent texts. 
The latter \textit{post-hoc} approaches are further structured into \textbf{interaction attribution} (\S \ref{sec:attribution}) tracing a prediction back to feature interactions between the model's two inputs, \textbf{global explanation} verifying the consistency of embeddings for known text relations on a dataset level and \textbf{surrogate modeling} (\S \ref{sec:surrogate}) optimizing a secondary interpretable model to approximate the original one. 
We begin every sub-category by introducing the common idea behind the described methods and conclude with opportunities they open up as well as remaining challenges.
Table \ref{tab:approach-classification} in the Appendix visualizes our taxonomy, links respective sections and points to code resources.
Finally, we examine evaluation methods and data sets (\S\ref{sec:evaluation}), and conclude with an extended discussion (\S \ref{sec:discussion}) highlighting trade-offs, lessons learned, challenges and future perspectives.

\section{Interpretable Embeddings} \label{sec:interpret_emb}

These approaches aim at structuring the embedding space so that it reflects human-understandable features. As such they create inherently interpretable models \citep{stop_explaining}.

\subsection{Shaping interpretable spaces} \label{sec:space-shaping}

\paragraph{Idea.}
An interpretable embedding space can be explicitly trained to express human-understandable aspects, thereby bridging the gap between the power of neural embeddings and the interpretability of classic methods based on ``bag-of-words'' representations.

\paragraph{QA features} aim to develop interpretable features by framing embedding generation as answering a set of predefined questions about a text and encoding the answers as features. For this, we first need to find \textit{a suitable set of questions} about texts, and create training data that \textit{elicits answers} to these questions. 
Specifically, \citet{benara2024crafting} let an LLM answer ``Yes''/``No'' questions about a text (\textit{Is the text about sports?} \textit{Does the text express a command?}), building prompts based on dataset description. For predicting fMRI responses to language stimuli their method outperforms several baselines. \citet{sun2024general} propose constructing a concept space from a dataset by clustering word embeddings and then applying two constraints. First, the QA prompts must be strongly associated with one of  the detected clusters. Second, for positive text pairs, all questions should be answered with ``Yes'' and for negative text pairs  with ``No'', to sharpen the boundary between similar and dissimilar texts.  
The resulting embeddings are interpretable in the sense that question answers can be directly inferred from them.

\begin{figure}[t]
    \centering
    \includegraphics[width=1\linewidth]{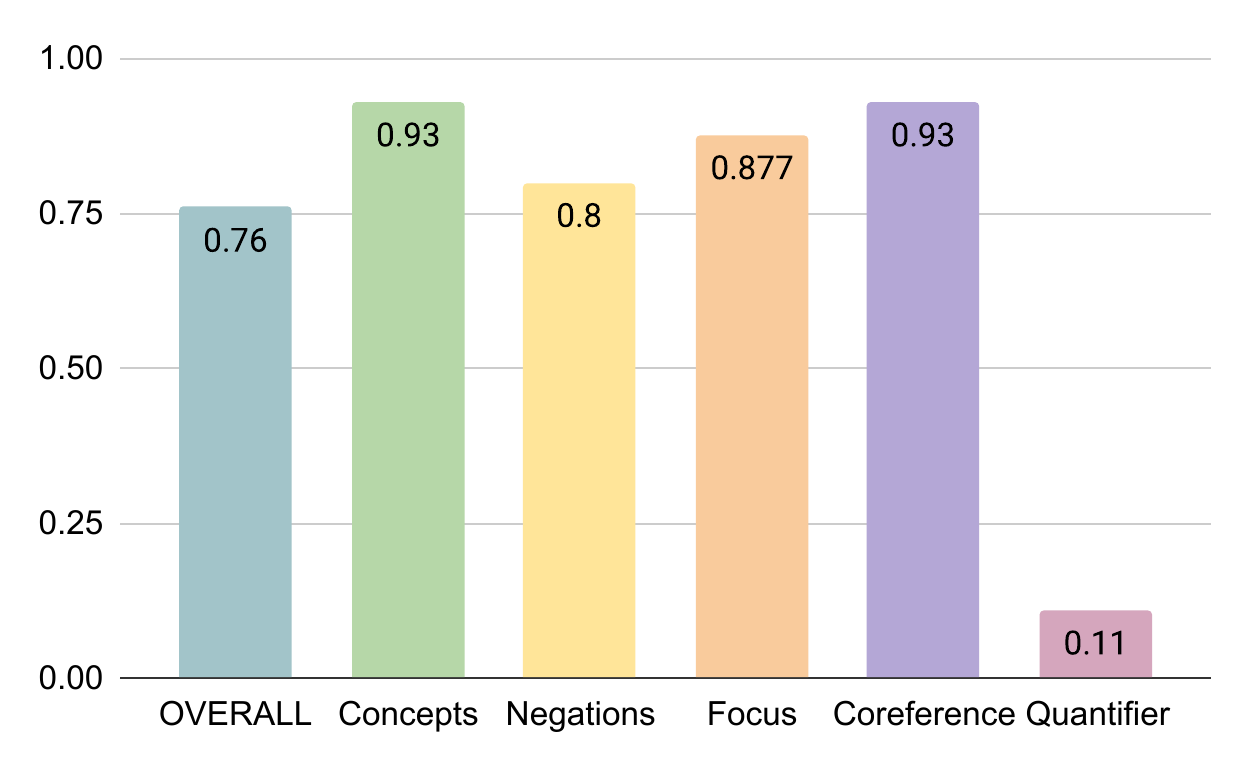}
    \caption{In S3BERT space decomposition, an overall $sim$=0.76 for the sentence pair \textit{Two men are singing} and \textit{Three men are singing}   
    emerges from aggregating per-aspect similarities. (Simplified aspect set used here.)}
    \label{fig:s3bertexample}
\end{figure}

\paragraph{Sub-embedding features.} An embedding space can be  decomposed into \textit{multidimensional subspaces}, each isolating a specific semantic aspect. This allows overall similarity to be broken down into aspect-specific scores.
The approach by \citet{opitz-frank-2022-sbert} (``S3BERT'') requires a user to define a set of metrics that measure interpretable similarity aspects of two texts (e.g., \textit{Is the focus of the texts the same?}) and binds those in respective sub-spaces. Since such aspects often are implicit in the texts, they leverage abstract meaning representation graphs \citep{banarescu-etal-2013-abstract} that encode aspects such as number, focus, semantic roles, negation; and use graph matching metrics \citep{opitz-2023-smatch} on aspectual subgraphs. They fine-tune a reference embedding model such that the similarity of aspectual sub-embeddings regresses to the aspectual graph metrics. A consistency loss and residual sub-embedding help tie the overall similarities to the original reference. 
In the example in Figure \ref{fig:s3bertexample}, the similarity of concepts increases the value, while the dissimilarity of quantificational structure correctly lowers it. 

Other approaches omit the consistency loss and aim to induce entirely new decomposed spaces. For instance, ``multi-facet'' embeddings are learned with graph metric ground truth \citep{risch-etal-2021-multifaceted},  or ``specialized-aspect'' embeddings with aspect-specific transformer encoders \citep{10.1145/3529372.3530912specialized, schopf-etal-2023-aspectcse}. 

A more coarse-grained decomposition is proposed by \citet{ponwitayarat-etal-2024-space}, whose linguistic analysis of the \textit{S}emantic \textit{T}extual \textit{S}imilarity dataset (\textit{STS}) \citep{stsb} found that a single continuous similarity range is not sufficient. They suggest a decomposition into two spaces, one for loosely similar texts (lower range), and another to capture finer distinctions among highly similar texts (higher range).

\paragraph{Anchor features} let individual embedding dimensions express association (i.e., similarity) to interpretable anchors in a database, e.g., representative prototype texts that have been sampled \citep{wang-etal-2025-ldir} or pre-computed and aligned topics \citep{10.1007/978-3-540-78646-7_51}. Compared to QA features, the provided explanation is more indirect but shows greater accuracy, almost on par with their non-interpretable embedding counterparts.

\paragraph{Challenges and opportunities.}
QA-based approaches have been evaluated favorably against bag-of-words baselines \citep{sun2024general} and in specific domains \citep{benara2024crafting}. They still struggle with matching the performance of reference embedding models, likely due to the difficulty of defining a general and complete set of questions.\\
Similarly, sub-embedding decomposition approaches require the manual definition of semantic aspects. However, the resulting dimensions are not directly interpretable as features---only the similarity values they produce can be linked to the defined aspects. This reliance on handcrafted features, can be seen as a limitation on the one hand, but enables the definition of custom embedding spaces aligned with specific interpretability goals on the other hand.\\
All space-shaping approaches pose additional constraints on a model risking downstream accuracy compared with standard embeddings.
Interestingly, sometimes they can induce regulatory effects. 
The S3BERT authors e.g. observed a significant performance increase for judging argument similarity.
\subsection{Sparse representations} \label{sec:sparse}

\paragraph{Idea.} Instead of assigning individual dimensions of dense embeddings to certain aspects, another approach towards creating interpretable spaces is sparsification. 

\paragraph{Unsupervised Sparsification.} 
Such sparsity can be induced without supervision by learning to reconstruct the embedding from sparse latent variables \citep{faruqui-etal-2015-sparse, prokhorov-etal-2021-learning, o2024disentangling}. \citet{trifonov-etal-2018-learning} find that such sparse embeddings can indeed isolate some dimensions corresponding to human-interpretable features, including even spatial object relations (e.g., \textit{physically laying on something}). However, they note that it can be difficult to tell ``which features a dimension captures'', and that the ``increase in interpretability comes at a cost in reconstruction quality and, in some cases, utility in downstream tasks.''

\paragraph{Sparse lexical embeddings} map input texts onto term weight vectors whose dimensionality equals the length of a predefined vocabulary.
Transformer models naturally provide this capability.
Applying the unembedding matrix to the last layer's token representations results in a logit vector over the model's vocabulary (the basis for masked or next token classification during pre-training).
Sparse embeddings repurpose them by combining the token-level logits into  text-level representations through pooling along the sequence dimension. A sparsification objective is applied during contrastive learning.
In contrast to lexical approaches like tf-idf, they are not bound to terms in the actual input but can assign weights to \textit{expansion terms} that may additionally be relevant in the given context, e.g. synonyms.
Sparse lexical embeddings are popular in retrieval scenarios because the term-based representation enables deployment via efficient inverted indices.
\citet{deepct} predict lexical term-weights from contextualized embeddings, \citet{sparterm, sparta} introduce vocabulary expansion, and \citet{splade, splade_v2} propose an end-to-end trainable model. Recently, sparse and dense embeddings have also been combined in unified models \cite{sparse_embed, zhang-etal-2024-mgte, granite}.

\paragraph{Challenges and opportunities.}
Unsupervised sparse features can correspond to intuitive text characteristics but can be difficult to interpret in other cases.
In turn, sparse lexical representations are trivial to interpret. Their sparsity can be beneficial in suitable scenarios like building a search index.
However, the need for specialized data structures in order to handle their high dimensionality efficiently may be a burden in other contexts.

\subsection{Structured Objects} \label{sec:geometric}

\paragraph{Idea.} Certain text relationships are inherently asymmetric. For instance, a natural relation between texts is \textit{entailment}: A given hypothesis follows from a premise. Geometric embeddings reach beyond vector representations and utilize structured objects for representation offering a way to model these relationships. 

\paragraph{Box embeddings} represent inputs as high-dimensional boxes. Assume two 2-dimensional boxes $a$ and $b$, centered at zero with the left lower corner. We have their size $s_a = a_1 \cdot a_2$, $s_b = b_1 \cdot b_2$, and their overlap $o_{a, b}=min(a_1, b_1) \cdot min(a_2, b_2)$. We can define their similarity as mutual containment, $o_{a,b} / (s_a + s_b -o_{a,b})$,
and quantify an asymmetric relationship like the entailment as unidirectional containment: $o_{a,b} / s_b$ is exactly $1$ if $b$ is fully contained/entailed in/by $a$. The challenge is to learn such objects given the 'curse of dimensionality', according to which box size and overlap tend towards zero for high-dimensional spaces. To alleviate such learning problems, \citet{chheda-etal-2021-box} propose to adopt a probabilistic soft box overlap formulation based on Gumbel random variables \citep{dasgupta2020improving}. 

\paragraph{Distributional embeddings} view a text as a random variable (RV). Intuitively, this provides us with a model of multiple interpretations, which seems appealing due to natural language ambiguity: A text can have multiple interpretations, and only some of these interpretations can map to those of another similar text. But how to build such a probabilistic space? \citet{shen-etal-2023-sen2pro} model a text as a Gaussian RV embedding $\mathcal{N}_d(\mu, \Sigma)$ by estimating ``Model uncertainty'' via Monte Carlo Dropout \citep{pmlr-v48-gal16}, and data uncertainty via smaller linguistic perturbations (e.g., dropping a word). The covariance matrix ($\hat{\Sigma}$) is then efficiently approximated through a banding estimator \citep{Bien02042016}. For increased efficiency, \citet{yoda-etal-2024-sentence} directly predict mean ($\hat{\mu}$) and covariance ($\hat{\Sigma}$).

\paragraph{Operator learning.}
An approach that works with standard vector representations but can also account for asymmetric text relations are the work by \citet{huang-etal-2023-bridging}.
They propose learning interpretable operators for text meaning composition, such as union or fusion. These operators are modeled using neural networks, and the embedding space is retrained to accommodate such operations. Their evaluation shows minimal loss in standard similarity tasks, but greatly improved performance for compositional generation tasks.

\paragraph{Challenges and Opportunities.} 
Modeling embeddings as geometric objects or learning operators can account for the long-established argument that similarity relations need not be symmetric \cite{tversky, lin}.
However, these approaches introduce additional complexity that may be too much of an overhead in other applications that do not have this requirement.

\subsection{Set-based Interpretability} \label{sec:setbased}

\paragraph{Idea.}
Set-based approaches are based on \textit{two sets} of embeddings rather than two points. Often sets consist of token embeddings (from the last layer of an encoder), but other approaches go further and build meta-sets of text embeddings. Aligning such sets can reveal how different parts of the texts relate and contribute to the overall similarity score. 

\paragraph{Token weight embeddings} build text representations by aggregating token-level embeddings with explicit weights that reflect each token's importance and provides insight into how individual tokens contribute to the final text embedding. 
E.g., \citet{wang-sbertwk} estimate token importance and novelty weights by analyzing variance across transformer layers. \citet{seo-token-attention}, train models to learn token weights directly, utilizing a reconstruction loss. 
\citet{minishlab2024model2vec} compute static embeddings for all vocabulary tokens via a single transformer forward pass per token, followed by Zipf-informed averaging.
As the final text representations are single vectors, while interpretability is on the level of individual tokens, these approaches are at the intersection between dense and set-based embeddings.

\begin{figure}
    \centering
    \includegraphics[width=\linewidth]{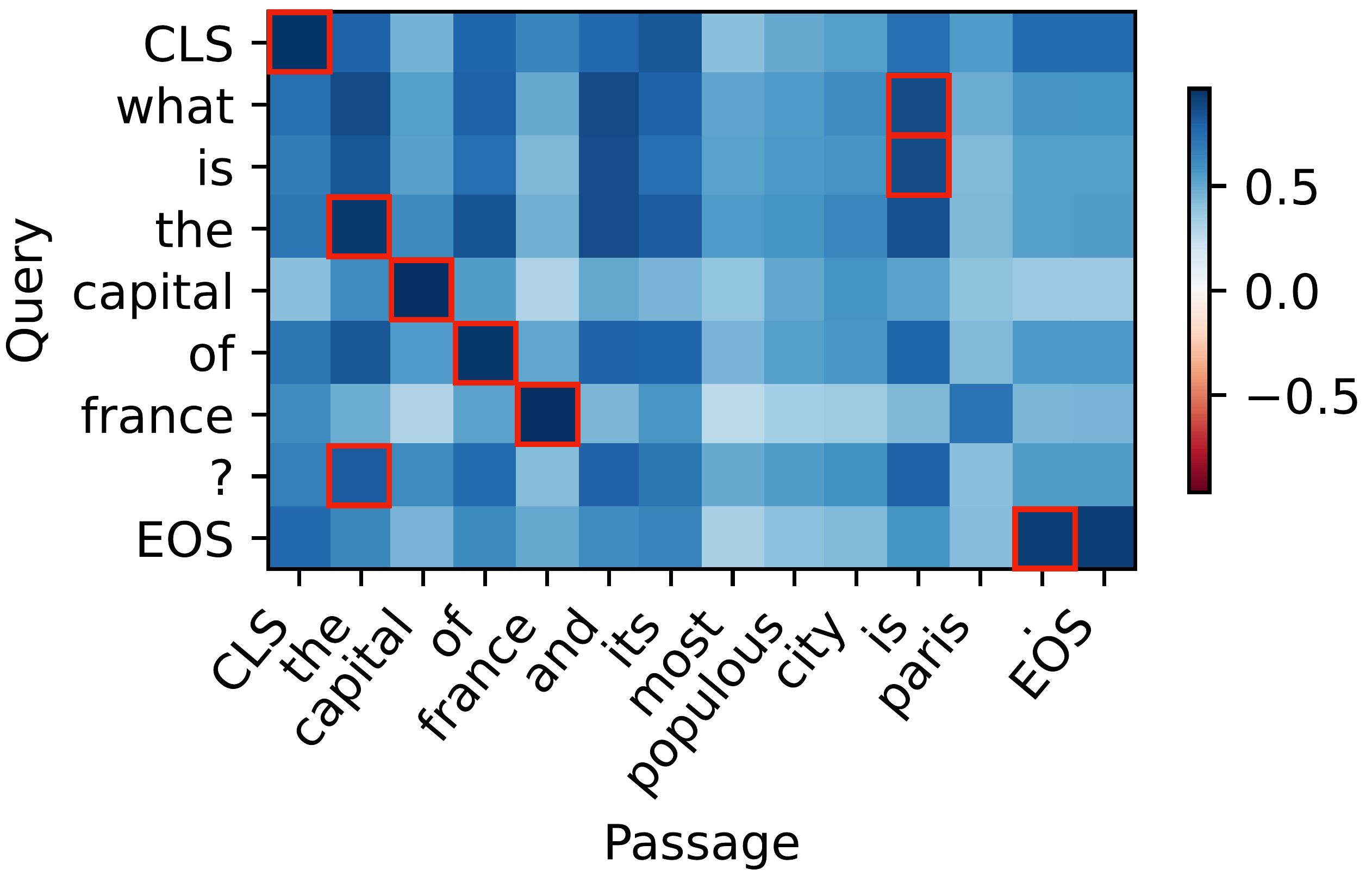}
    \caption{An example of a late-interaction matrix between query and passage token embeddings in the ColBERTv2.0 model. The overall $sim$ is 0.965. Red boxes indicate row-wise maxima (alignment).}
    \label{fig:colbert}
\end{figure}

\paragraph{Sequential embeddings} compare embeddings from the final model layer---before any reduction (``late interaction''). Two prominent techniques are ColBERT and BERTscore \citep{10.1145/3397271.3401075Colbert, santhanam-etal-2022-colbertv2}, both of which compute asymmetric max-alignments between tokens and aggregate the similarities of the aligned pairs.  BERTscore performs this alignment in both directions to produce a symmetric similarity measure. \\
In terms of \textit{explainability}, both approaches derive the final similarity score from token-level alignments, showing approximately which tokens the model matches between the inputs (Figure \ref{fig:colbert}). 

\paragraph{Multi-view Interpretation.}

Some approaches extend this idea by generating multiple text-level embeddings, each reflecting a different view or interpretation of the input text.

\citet{hoyle-etal-2023-natural} use a generative model to produce alternative hypotheses about a text.
\citep{ravfogel2024descriptionbased} decompose the text into smaller statements or descriptions.
Given a decomposition of a text $x$ into smaller parts $\{x_1, ...x_n\}$, the embedding model is then applied to each part individually, producing a set of text embeddings $\{e_1, ...e_n\}$. 

A variation of the multi-text set-based approach is proposed by  \citet{liu2024meaning}. To compute textual similarity, they sample sets of possible continuation from an LLM and calculate the average log-likelihood difference between each input text and the generated continuations. The continuations can then be examined to provide an interpretable basis for the resulting similarity score.

Finally, \citet{liu2024conjuring} compute text similarity \textit{multi-modally} by comparing the imagery evoked by each text, using denoising via Stochastic Differential Equations \citep{song2021scorebased}. Similarity is higher when texts elicit similar images, enabling visual interpretation of the score.

\subsection{Challenges and Opportunities}

Set-based approaches enable interpretable alignment of token-level embeddings, which can be valuable for tasks such as identifying semantic differences between related documents \citep{vamvas-sennrich-2023-towards}. They also naturally support asymmetric text relationships via directional matching or alignment. 
An important limitation is that sets of embeddings typically consume more memory than single vectors. 
Sequential embeddings also do not have a fixed size but vary with input length.
While decomposition based approaches can point out matching sub-statements or hypotheses, they can require multiple forward passes.

\section{Post-hoc Explanation} \label{sec:post-hoc}

Different from inherently interpretable models, post-hoc explanations employ an additional method to gain insights into the predictions of a black-box model.

\subsection{Interaction Attribution} \label{sec:attribution}

\paragraph{Idea.}
Attribution-based approaches aim at assigning importance values to features reflecting their contribution to a given prediction of a model.
A special characteristic of similarity models is that their predictions do not depend on individual features, due to the multiplicative interaction between the two inputs’ embeddings in $sim$. 
Attributions must, therefore, be to feature interactions. 
First-order methods do not suffice to explain such interaction \cite{shapley_interaction, int_hessians}, and second-order methods are required for attribution in similarity models.
Two lines of work have addressed this issue in text similarity models.

\begin{figure}
    \centering
    \includegraphics[width=\linewidth]{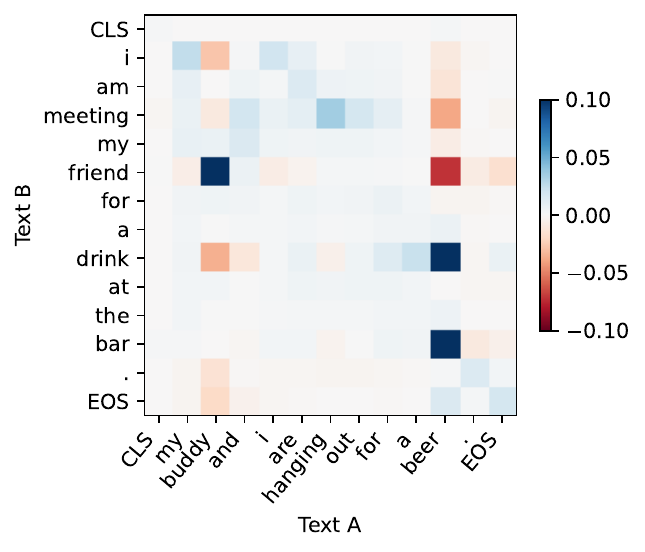}
    \caption{Interaction attributions between two sentences computed with the IJ method.
    The $sim$ is 0.618 and the measurable attribution error is 0.001.}
    \label{fig:attribution-example}
\end{figure}

\paragraph{Integrated Jacobians.}
Integrated gradients (IG) attributes a scalar model prediction back onto individual input features by integrating over a number of interpolations between the actual input and an uninformative reference input \citep{ig}. \citet{moeller-etal-2023-attribution} have applied the underlying theory of IG to text embedding models and proposed \textit{Integrated Jacobians} (IJ) as the equivalent of IG for this model class.
For text embedding models the output takes the form of a token-token matrix, showing the contribution of all individual token interactions to the $sim$ (Figure \ref{fig:attribution-example}). An approximate version of these attributions is directly applicable to off-the-shelf models without a need for tuning \citep{moeller-etal-2024-approximate}.

\paragraph{Relevance Propagation.}
Layer-wise relevance propagation (LRP) is a framework to propagate feature-importance values for a model prediction back through the model in a layer-wise fashion \cite{lrp, lrp_survey}. Propagation rules are derived for individual layers based on first-order Taylor expansion of the underlying function. BiLRP extends the LRP framework to Siamese similarity models. Similar to IJ, the computation also takes the form of a product between two Jacobian-like matrices. The method was originally proposed in the computer vision domain \cite{bilrp} and has recently also been applied to Siamese text encoder models \citep{vasileiou-eberle-2024-explaining}.

\paragraph{Challenges and Opportunities.}
Attribution approaches need to build Jacobian matrices, coming at a temporal complexity of \(2\!\times\! d\)
independent backward passes, \(d\) being the model's embedding dimensionality. The resulting Jacobians have a quadratic spatial complexity 
and can require large GPUs to compute the associated matrix multiplications efficiently.
Despite the computational costs, attribution methods have the advantage of being applicable to a wide class of embedding models as long as they are differentiable.
They can provide certain theoretical guarantees \citep{ig, int_hessians}, but have also been proven to be subject to other fundamental limitations \citep{bilodeau}.

\subsection{Global explainability} \label{sec:global}

\paragraph{Idea.}
A common way of differentiating explainability methods is into \textit{local} and \textit{global} explanation \citep{danilevsky-etal-2020-survey}. 
Local approaches work on the level of individual examples.
Alternatively, we can globally analyze the geometry of embeddings using a dedicated evaluation dataset. 

\paragraph{Text relations.}
\citet{Zhu18} and \citet{Zhu20} follow this approach by constructing sets of sentences with known relations based on linguistic properties.
In their initial work, the authors use triplets of sentences including a pair known to be similar with regard to a certain property and a dissimilar negative.
Properties include negation, passivation, change of syntactic roles and word-ordering.
It is then evaluated how consistently positive pairs are closer to another in the representation space than to the negative. 
In the second publication, the group extends the analysis to quadruples consisting of two pairs of similar sentences, thus, evaluating the similarity of sentence relations. 

\paragraph{Challenges and opportunities.}
Analyzing embedding geometry provides a higher-level understanding of how consistently relations between sentences are represented in an embedding space. 
However, it requires the manual construction of suitable evaluation sets targeting specific properties and insights are limited to the properties covered.

\subsection{Surrogate modeling} \label{sec:surrogate}

\paragraph{Idea.}
Surrogate models approximate a complex black-box model with a simpler, typically linear model that is inherently interpretable.
We differentiate between two types, models operating on interpretable features approximating the original models predictions and models operating on the original model's embeddings \textit{probing} them for known properties.

\paragraph{Interpretable approximation.}
\citet{nikolaev-pado-2023-representation} construct artificial sentence pairs with known linguistic features.
Based on these features, they then fit surrogate regression models to match the cosine similarity scores of different sentence transformers.
The linearity of the fitted surrogate models allows them to analyze the relative importance of linguistic aspects in sentence pairs through the weights assigned to respective features. 

\paragraph{Probing.}
A \textit{probe} is a (often linear) classification model that is trained on top of pre-trained sentence embedding for a defined task.
It assesses the generalizability of an embedding by testing whether the associated property is (linearly) separable in the learned representation space.
\citet{Conneau18} propose ten tasks around surface-level, syntactic and semantic information to probe the linguistic information contained in sentence representations of different models.
In another work the group tests applicability to downstream applications like sentiment classification or retrieval \citep{Conneau17}.
More recently, probing has become an important evaluation tool in large-scale text embedding benchmarks like MTEB \citep{muennighoff-etal-2023-mteb}.
\citet{nikolaev-pado-2023-investigating} investigate which layers in different models encode semantic information through probing. \citet{tehenan2025mechanistic}, inspired from ideas of mechanistic interpretability \citep{bricken2023monosemanticity}, use sparse dictionary learning for investigating token-level linguistic information that is pooled in a sentence embedding.

\paragraph{Challenges and Opportunities.}
While conceptually simple, surrogate models do require additional objectives and optimization to generate insights into black-box models.
Although limited, they do have a certain learning capacity, which needs not necessarily align with what the original model has learned.

\section{Evaluation and Datasets}
\label{sec:evaluation}

\subsection{Evaluation}

The presented approaches  differ substantially in the types of explanations they produce, making it difficult---if not impossible---to define a unified evaluation framework covering them all. In fact, evaluation often focuses on specific characteristics of individual approaches. Space-shaping models can explicitly correlate ground truth values for similarity metrics against predictions from respective sub-spaces \citep{opitz-frank-2022-sbert}. Aspect encoders test in how far nearest neighbors in the embedding space share aspects and assess whether aspect clusters emerge in dimensionality reduction plots \citep{10.1145/3529372.3530912specialized, schopf-etal-2023-aspectcse}. Specialized and structured objects allow to evaluate whether the learned representations reflect asymmetric relations like entailment or noun-hierarchy in WordNet \citep{yoda-etal-2024-sentence, chheda-etal-2021-box}. A typical procedure in the attributions field is iterative insertion or deletion of the most attributed features and simultaneous reevaluation of the predicted similarity between these perturbed inputs \citep{vasileiou-eberle-2024-explaining}. If the most attributed features are indeed the most important, the prediction should change drastically upon their perturbation. Sparse representations that are induced in an unsupervised way can be assessed through topic-coherence measures \citep{trifonov-etal-2018-learning}.

A central challenge in  evaluating explainability and interpretability is the absence of ground truth  for what constitutes a correct or valid explanation. \citet{vasileiou-eberle-2024-explaining} address this  by generating synthetic data using a rule-based similarity model. More commonly,  evaluation focuses on  performance trade-offs between  interpretable methods and their standard counterparts. The explanation quality  is  often  assessed qualitatively through example-based analysis. Some studies employ human evaluation to obtain  subjective judgements of explanation quality. However, this introduces additional parameters to the evaluation scenario, e.g., which target group an explanation method aims at \citep{8920711-targetgroup}.

\subsection{Datasets}\label{ssec:datasets}

Datasets can serve at least \textit{two purposes} within the realm of interpretable embeddings and semantic search explanations. The first purpose is a potential application to \textit{evaluate a method's explanation} against human explanation. The second purpose is \textit{global explanation} through evaluating embedding models on text pairs with controlled relation. 

\paragraph{Human Explanations.} \citet{LOPEZGAZPIO2017186} release the i(nterpretable)STS data set that elicits relations and similarities between individual segments of texts. \citet{deshpande-etal-2023-c} propose the C(onditional)STS dataset that elicits similarity values for specific aspect of interest. The theory that underlies iSTS aligns with attribution or set-based approaches, while CSTS is motivated by a more abstract multi-aspect view akin to what is sought by feature-based explainability methods. 

\paragraph{Interpretable Text Relations.}
For their analysis \citet{Zhu18} and \citet{Zhu20} construct two datasets of sentence triplets and quadruples consisting of a negative and positive pairs with a shared linguistic property. 
\citet{li2025sentence} propose a neuro-symbolic tool for automatically creating such sets and use the resulting data for ranking text embedding models in interpretable linguistic categories. The STS3k data set \citep{fodor2024compositionality} contains sentence pairs with systematic word combinations, rated for semantic similarity by human participants. \citet{nastase-merlo-2024-tracking} propose specialized sentence sets to study the grammatical information that resides in an embedding.

\section{Discussion}
\label{sec:discussion}

\paragraph{Method trade-offs.} The surveyed methods differ in their conceptualization of interpretability, computational cost, fidelity to input tokens, and dependence on specific model architectures. All variants of \textbf{interpretable embeddings} produce inherently interpretable models, offering transparency into their decision-making processes by design \cite{stop_explaining}. 
However, they often pose additional constraints on models which can lead to compromises in predictive performance.
In contrast, \textbf{post-hoc methods} do not constrain models upon training but rely on additional computation, surrogate optimization and specialized datasets to generate insights.

\paragraph{What is the ``right'' explanation?} Given the above trade-offs none of the presented approaches should be seen as to provide true, unique and faithful explanations \cite{murdoch}. At the same time, all of them provide insights into text embedding models going beyond a black-box embedding or single scalar similarity score. Each approach may lead to hypotheses about where these models fail and how they can be improved \citep{wiegreffe}. Rather than competing for a single best explanation, therefore, we suggest to consider individual methods as independent pieces of evidence.

\paragraph{Lessons learned.} Following the above argumentation, we can already draw a number of overarching conclusions about text embedding models: 
They encode a wide range of linguistic knowledge, including syntax and semantic information like tense \cite{conneau-kiela-2018-senteval, huang-etal-2021-disentangling}, learn to match synonyms well \cite{moeller-etal-2024-approximate, Zhu18} (cf. Fig. \ref{fig:attribution-example}) and successfully ignore irrelevant parts of sentences \citep{nikolaev-pado-2023-representation}.
But they often do not sufficiently account for negation or random word deletion \citep{weller-etal-2024-nevir, Zhu20}. 
They largely rely on nouns and verbs \citep{vasileiou-eberle-2024-explaining, nikolaev-pado-2023-representation} as well as subjects, predicates and objects \cite{moeller-etal-2024-approximate}. 
Nevertheless, they do require the full contexts of sentences for their predictions to be reliable \citep{moeller-etal-2023-attribution} and are sensitive to word order \citep{Zhu18}.\\
With space-shaping methods, we have the ability to actively manipulate encoded information \citep{sun2024general, shen-etal-2023-sen2pro, schopf-etal-2023-aspectcse} and can e.g. correct embeddings to account for negation \citep{opitz-frank-2022-sbert}. Finally, structured embeddings have proven to successfully account for non-symmetric text relations \citep{yoda-etal-2024-sentence, dasgupta2020improving}.

\paragraph{Upcoming challenges.} % Wie wäre "Open challenges"?
As models become capable of ingesting longer context \cite{zhang-etal-2024-mgte, xiong-etal-2024-effective}, we may wonder if interpretability approaches transfer to explaining the similarity of \textbf{long documents}. 
Fine-grained explanations such as token attributions or alignments may require an aggregation step to a sentence or paragraph level balancing higher-level interpretability and compute scaling.

The embedding research landscape has also found another recent focus in \textbf{multilinguality} \citep{wang2024multilingual}. 
It will be interesting to investigate cross-lingual text similarity and
we see an interesting tension here between capturing universal and language specific or cultural patterns.  

Embedding models are also increasingly used as parts in \textbf{more complex models} like Retrieval Augmented Generation \citep[RAG,][]{NEURIPS2020_6b493230_rag}.
The approaches presented here may be used to explain the retrieval step and interpretability approaches for generative models may be utilized to understand the compilation of responses (\cite{attnlrp}).
However, it is an open question how to combine explanations for the two steps.

The social-sciences and sensitive fields like legal text processing often work with text representations but come with explainability requirements.
A lack of interpretability can be a reason not to use state-of-the-art approaches in these fields, but to fall back onto outdated alternatives like simple dictionary-based approaches.
\textbf{Interdisciplinary efforts} should focus more on understanding and addressing these requirements.

Finally, the evident link between text similarity and embedding models motivates a closer look at \textbf{the notion of similarity}.
Similarity is known to be context-dependent \citep{Gardenfors, bar}, possibly asymmetric \citep{tversky}, and even intransitive \citep{lin-similarity}.
However, common datasets and benchmarks assume a one-dimensional scale and unification across various tasks and objectives \cite{stsb, gooaq, muennighoff-etal-2023-mteb}, which has enabled scalability but may be overly simplifying.
Evidence for this can be seen in the fact that successful attempts now often use instructions \cite{allen} or specialized adapters \cite{jina2} to condition a model for specific tasks (i.e. contexts).
This calls for interpretability research to better understand relevant text aspects in different contexts.

\paragraph{Perspective.}
Despite the additional challenges associated with the explainability of similarity, this survey has shown that there have been substantial efforts towards better understanding text embeddings and text similarity models.
Most of these approaches are directly applicable to the next generation embedding models derived from decoders, since after removing causal attention masking, adding pooling and contrastive training, their architecture is identical to the previous encoder-based generation.
This holds true for inherently interpretable embeddings, e.g. space-shaping and also post-hoc explanations, e.g. attribution whose only requirement is for models to be differentiable.
Challenges are not fundamental but on the level of implementation details and computational costs due to the larger number of parameters and embedding dimensions.
We believe an increased attention towards explainability research in this area can help not only understand and explain their outputs, but also mitigate biases and errors in these models, yielding improvement towards accuracy and safety.

\section*{Limitations}

Capturing the full breadth of the area of interpretable text embeddings and their similarity cost us some depth and exactness. For instance, in Section \ref{sec:stage}, we suggest to speak of \textit{interpretable} models when their predictions are inherently understandable by humans and define \textit{explanation} as a post-hoc process in contrast.
However, throughout the paper, we also use these terms interchangeably.

While we have put a lot of effort into identifying relevant publications for the scope of this survey, there is a chance that we missed some works.
Hence we suggest viewing our survey as a guide and introduction to this field that is representative, but possibly not fully exhaustive. 

Finally, interpretability research tends to lag behind the rapid evolution of state-of-the-art models. Most of the approaches we survey here built up on, or were applied to now outdated text embedding models. 
Nevertheless, these methods are often general enough to be transferred to state-of-the-art models.

\section*{Acknowledgments}

Three authors received funding through the project \textit{Impresso – Media Monitoring of the Past II Beyond Borders: Connecting Historical Newspapers and Radio}. Impresso is a research project funded by the Swiss National Science Foundation (SNSF 213585) and the Luxembourg National Research Fund (17498891).

\bibliography{custom}

\appendix

\section{Overview} 
\label{app:1}

\begin{table*}[ht]
    \centering
        \caption{A summary of our taxonomy with links to respective sections, publications and their code if available. The table is split into the two families of methods that are further divided into \textit{Types}, corresponding to subsections and \textit{Subtypes} corresponding to paragraphs in the main text. \textit{Train} is whether the method requires training and \textit{Approx.} refers to whether a method approximates the similarity score of a reference embedding model. When code is labeled as `NA', this means that we were not able to find a public code repository.}
    \label{tab:approach-classification}
    \adjustbox{width=\linewidth}{\begin{tabular}{llllll}
    \toprule
        Type & Subtype & Paper & Train & Approx.  &   code\\
        \midrule
        \multicolumn{6}{c}{\textbf{Interpretable Embeddings (\S\ref{sec:interpret_emb})}}\\
        \midrule
        \multirow{9}{*}{\textbf{space-shaping (\S\ref{sec:space-shaping})}} & \multirow{2}{*}{QA-features} & \citet{sun2024general} &  yes  & no & \href{https://github.com/dukesun99/CQG-MBQA}{github} \\
        & & \citet{benara2024crafting} & yes  & no & \href{https://github.com/csinva/interpretable-embeddings}{github} \\
        \cmidrule(r){3-6}
        & \multirow{5}{*}{sub-embedding} & \citet{opitz-frank-2022-sbert} & yes  & yes  &  \href{https://github.com/flipz357/S3BERT}{github} \\
        & & \citet{risch-etal-2021-multifaceted} & yes & no & \href{https://github.com/philipphager/faceted-domain-encoder}{github}\\
        & & \citet{10.1145/3529372.3530912specialized} & yes & no & \href{https://github.com/malteos/aspect-document-embeddings}{github} \\
        & & \citet{schopf-etal-2023-aspectcse} & yes & no & NA\\
        & & \citet{ponwitayarat-etal-2024-space} & yes & no &  \href{https://github.com/KornWtp/MixSP}{github} \\
        \cmidrule(r){3-6}
        & \multirow{2}{*}{anchors} & \citet{10.1007/978-3-540-78646-7_51} & yes & no & NA \\
        & & \citet{wang-etal-2025-ldir} & no & no & \href{https://github.com/szu-tera/LDIR}{github}\\
        \cmidrule(r){2-6}
        \multirow{6}{*}{\textbf{sparsification (\S\ref{sec:sparse})}} & \multirow{2}{*}{unsupervised} & \citet{trifonov-etal-2018-learning} & yes & yes & NA \\
        & & \citet{prokhorov-etal-2021-learning} & yes & yes  & \href{https://github.com/VictorProkhorov/HSVAE}
        {github}\\
        \cmidrule(r){3-6}
        & \multirow{4}{*}{lexical} & \citet{deepct} & yes & no & \href{https://github.com/AdeDZY/DeepCT}{github}\\
        & & \citet{sparterm} & yes & no & NA \\
        & & \citet{sparta} & yes & no & NA \\
        & & \citet{splade, splade_v2} & yes & no & \href{https://github.com/naver/splade}{github} \\
        \cmidrule(r){2-6}
        \multirow{5}{*}{\textbf{structured objects (\S\ref{sec:geometric})}} & \multirow{2}{*}{box embeddings}  & \citet{chheda-etal-2021-box} & yes & no & \href{https://github.com/iesl/box-embeddings}{github} \\
        & & \citet{dasgupta2020improving} & yes & no & \href{https://github.com/iesl/gumbel-box-embeddings}{github} \\
        \cmidrule(r){3-6}
        & \multirow{2}{*}{gaussian embeddings} & \citet{shen-etal-2023-sen2pro} & no & yes & NA \\
        & & \citet{yoda-etal-2024-sentence} & yes & no & \href{https://github.com/yoda122/GaussCSE}{github} \\
        \cmidrule(r){3-6}
        & operator learning & \citet{huang-etal-2023-bridging} & yes & yes & \href{https://github.com/jyhuang36/InterSent}{github}\\
        \cmidrule(r){2-6}
        \multirow{9}{*}{\textbf{set-based (\S\ref{sec:setbased})}} & \multirow{3}{*}{token-weights} & \citet{wang-sbertwk} & no & no & \href{https://github.com/BinWang28/SBERT-WK-Sentence-Embedding}{github}\\
        & & \citet{seo-token-attention} & yes & no & NA \\
        & & \citet{minishlab2024model2vec} & no & yes & \href{https://github.com/MinishLab/model2vec}{github} \\
        \cmidrule(r){3-6}
        & \multirow{2}{*}{sequential} & \citet{10.1145/3397271.3401075Colbert} & yes & yes  & \href{https://github.com/stanford-futuredata/ColBERT}{github} \\
         & & \citet{santhanam-etal-2022-colbertv2} & no & yes  & \href{https://github.com/stanford-futuredata/ColBERT}{github} \\
        \cmidrule(r){3-6}
        & \multirow{3}{*}{multi-view} & \citet{hoyle-etal-2023-natural} & no & no  & \href{https://github.com/ahoho/inferential-decompositions}{github} \\
        & & \citet{ravfogel2024descriptionbased} & no & no &  \href{https://github.com/shauli-ravfogel/descriptions}{github}\\
        & & \citet{liu2024meaning} & no & no & \href{https://github.com/tianyu139/meaning-as-trajectories}{github}\\
        \cmidrule(r){3-6}
        & image-set & \citet{liu2024conjuring} & yes & no & NA \\
        \midrule
        \multicolumn{6}{c}{\textbf{Post-hoc Explanation (\S\ref{sec:post-hoc})}}\\
        \midrule
        \multirow{2}{*}{\textbf{attribution (\S\ref{sec:attribution})}} & integrated Jacobians & \citet{moeller-etal-2023-attribution, moeller-etal-2024-approximate} & no & yes &  \href{https://github.com/lucasmllr/xsbert}{github} \\
        \cmidrule(r){3-6}
        & relevance propagation & \citet{vasileiou-eberle-2024-explaining} & no & yes &  \href{https://github.com/alevas/xai_similarity_transformers}{github} \\
        \cmidrule(r){2-6}
        \multirow{2}{*}{\textbf{global explanation (\S\ref{sec:global})}} & \multirow{2}{*}{text relations} & \citet{Zhu18} & no & no & NA \\
        & & \citet{Zhu20} & no & no & NA \\
        \cmidrule(r){2-6}
        \multirow{6}{*}{\textbf{surrogate modeling (\S\ref{sec:surrogate})}} & \multirow{1}{*}{interpretable approximation} & \citet{nikolaev-pado-2023-representation} & yes & yes & NA \\
        \cmidrule(r){3-6}
        & \multirow{4}{*}{probing} & \citet{Conneau17} & yes & no & \href{https://github.com/facebookresearch/InferSent}{github}\\
        & & \citet{Conneau18} & yes & no & \href{https://github.com/facebookresearch/SentEval/tree/main/data/probing}{github}\\
        & & \citet{nikolaev-pado-2023-investigating} & yes & yes & \href{https://github.com/macleginn/semantic-subspaces-code}{github} \\
        & & \citet{tehenan2025mechanistic} & yes & yes & \href{https://github.com/matthieu-perso/mechanistic_decomposition_sentences}{github} \\
        \bottomrule
    \end{tabular}}
    \end{table*}

\end{document}